\documentclass[graybox, natbib, nosecnum, twocolumn]{svmult}
\bibpunct{(}{)}{;}{a}{}{,} 

\usepackage{mathptmx}       
\usepackage{helvet}         
\usepackage{courier}        
\usepackage{type1cm}        

\usepackage{makeidx}         
\usepackage{graphicx}        
\usepackage{multicol}        
\usepackage[bottom]{footmisc}
\usepackage[normalem]{ulem}	
\usepackage{hyperref}  
\usepackage{soul}   
\usepackage[per-mode=symbol]{siunitx}

\usepackage[printonlyused]{acronym}

\acrodef{GPS}{Global Positioning System}
\acrodef{UAV}{Unmanned Aerial Vehicle}
\acrodef{UGV}{Unmanned Ground Vehicle}

\begin{document}

\title*{Robotics in Snow and Ice}
\author{François Pomerleau}
\institute{François Pomerleau \at Laval University, Québec, Canada, \email{francois.pomerleau@norlab.ulaval.ca}}
%
%
\maketitle
\section{Synonyms}
Robotics in the Arctic; 
Robotics in Antarctica;
Robotics in polar regions; 
Robotics in subarctic areas; 
Robotics in the cryosphere;
Robotics in cold environments.

\section{Definitions}
The terms \emph{robotics in snow and ice} refers to robotic systems being studied, developed, and used in areas where water can be found in its solid state.
This specialized branch of field robotics investigates the impact of extreme conditions related to cold environments on autonomous vehicles.

\section{Overview}

Regions of Earth where water can be found in its solid form are within the realm of the cryosphere.
As these regions required humans to be protected against the cold, which reduces mobility and operational time, robots have been developed to collect data or act remotely on the environment on their behalf.
Robotic systems deployed in the cryosphere need to be robust to subzero temperature while being challenged by a large spectrum of precipitation and terrain.
On the ground, sub-domains of the cryosphere include snow cover, freshwater ice, frozen ground, sea ice, glaciers, ice caps, and ice sheet.
In the air, solid water can take the form of freezing rain, hail, pellets, snow, and everything in between.
The combination of air temperature, precipitation, and ground conditions makes it very challenging for robotic systems to accomplish a given task as their visibility, maneuverability, and operation time will be impacted.
Broadly speaking, these conditions have pushed forward the research of novel solutions related to perception algorithms, control algorithms, locomotion designs, physical integrity, and the energy management of robots.
Originally a testbed for space exploration, applications extended rapidly to fundamental Earth sciences, transportation, and forestry.
While research related to \emph{robotics in snow and ice} is still in its infancy, it is a clear fertile ground for discoveries related to remote locations and the robustness of robots.

\subsection{Impact of the cryosphere on robotics}

\begin{figure*}[htbp]
\begin{center}
\includegraphics[width=\textwidth,trim={2px 0 0 3px},clip]{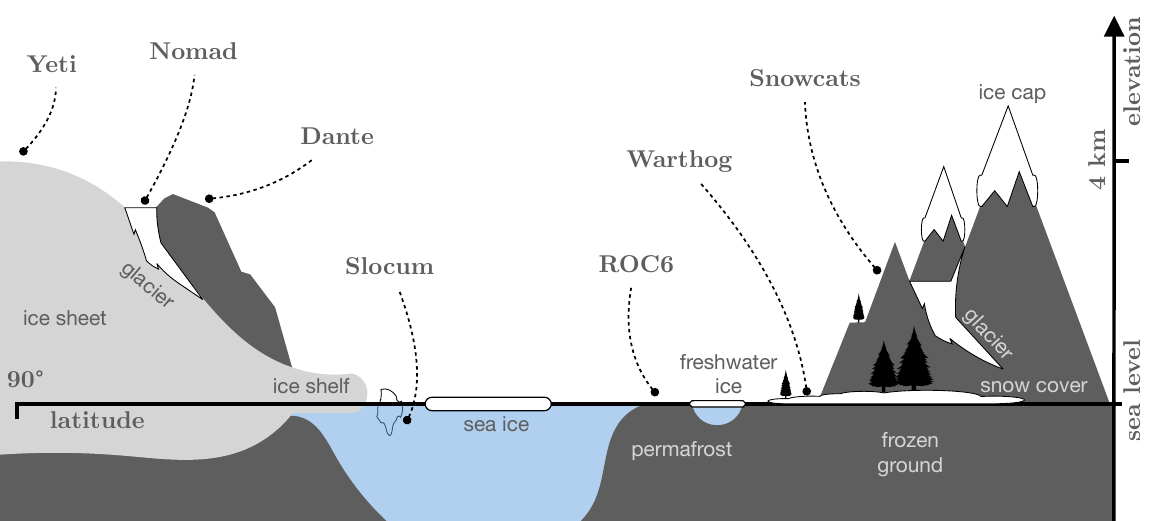}
\caption{Representation of sub-domains of the cryosphere as a function of latitude and elevation. 
Light blue represents water and labels with dashed lines highlight where key robots were deployed in different sub-domains.
The size of vegetation goes down with latitude and elevation.
}
\label{fig:zones}
\end{center}
\end{figure*}

Robotic systems must be well adapted for the specificities of locations where they need to achieve a task.
Cold environments are often categorized by their latitudes (e.g., Arctic, Antarctic, polar regions, subarctic) and elevation.
Instead of these two axes, we will be using directly sub-domains of the cryosphere to explain key elements of \emph{robotics in snow and ice}.
\autoref{fig:zones} relates latitudes and elevation to these sub-domains while highlighting key robots that were deployed in the cryosphere.
The historical context in which these robots were deployed will be described in later sections.

\begin{figure}[hbtp]
\begin{center}
\includegraphics[width=\columnwidth]{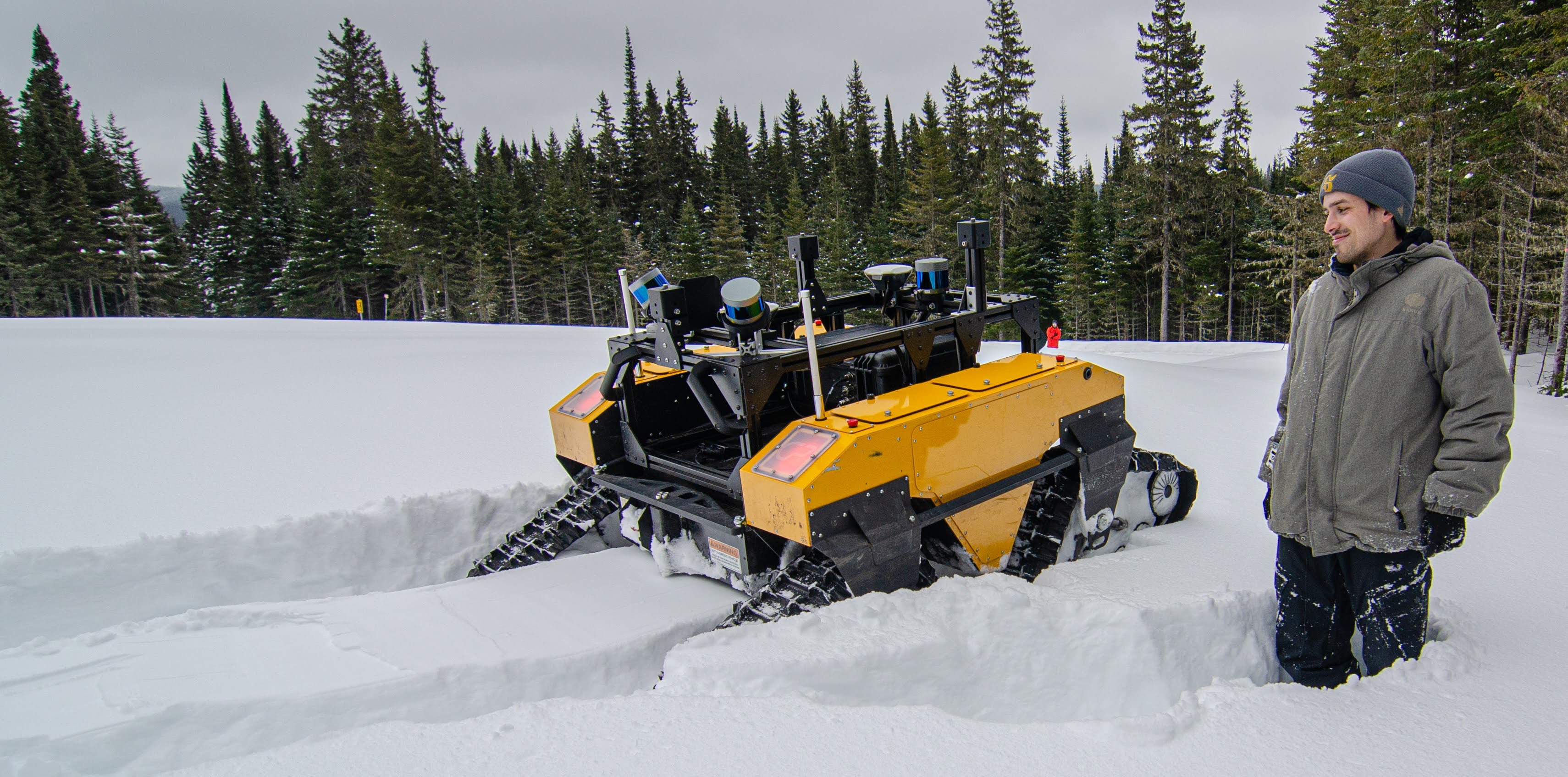}
\caption{A \SI{500}{\kg} robot navigating in a subarctic region. 
Specialized tracks allow the robot to navigate over deep and soft snow.}
\label{fig:deep_snow}
\end{center}
\end{figure}

The most common sub-domain of the cryosphere is snow covers.
They happen in any place where snow has time to accumulate onto the ground before melting.
The texture of snow cover is complex and highly varied (e.g., fresh powder, layered ice crusts, and compacted by winds) and constantly evolving.
Navigation through fresh snowfield is challenging due to deep sinkage, snow resistance, traction loss, and ingestion of snow into the drive mechanism~\citep{Lever2009}.
As shown in \autoref{fig:deep_snow}, specialized designs must be considered to ensure that a vehicle can rise over a snowbank when driving off-road.
Even with specialized locomotion designs, tight turns in soft snow can result in traction loss and wheel sinkage causing the robot to stall~\citep{Stansbury2004}. 
On the perception side, snow covers are highly reflective and produce few textures, which is challenging for vision-based algorithms.
For example, early field deployments in Antarctica could not rely on stereo vision because of missing texture causing disparity matching to fail~\citep{Moorehead1999}. 
Moreover, camera autoexposure algorithms tend to struggle on sunny days as the luminosity extends beyond the typical dynamic range of the photosensor, the texture used for feature extraction in localization algorithms is washed out, the ground constantly changes with winds, and images using auto-white balance will shift toward the blue spectrum~\citep{Paton2017}.
Even when specialized algorithms are used to enhance images, the number of features will be significantly lower than that of a typical outdoor scene~\citep{Williams2009}.

Another sub-domain to consider is waterbodies, whether they are fresh or salty, as they have unusual dynamics that robotic systems must be aware of.
For example, water around \SI{4}{\celsius} is denser than the water around \SI{0}{\celsius}.
This density variation results in shallow ponds freezing completely, while lakes and rivers will produce a large frozen crust protecting liquid water maintained at around \SI{4}{\celsius}.
A key application affected by this phenomenon is good transportation in northern regions.
This kind of transport relies on ice roads, which are mostly crossing frozen lakes and rivers.
Advancement in autonomous transportation has the potential of reducing transportation risks, but robots must be conceived with an understanding of ice formation.
Sea ice follows the same process, but the crust will tend to break under external forces, such as tides and winds. 
This sea ice must not be misclassified with icebergs, which also float on oceans, but were generally part of a large ice shelf that broke before drifting away.
The density of icebergs is so high that even icebreakers must avoid them during navigation. 
Moreover, icebergs are an important threat to permanent installations on water and to ship navigation in polar regions~\citep{Zhou2019}.
Any long-range autonomous navigation on water would need to monitor icebergs to avoid them.

Progressing toward the poles, we find more sub-domains of the cryosphere, such as ice shelves and glaciers, with the most prominent being ice sheets.
Greenland and Antarctica are the only ice sheets existing on Earth.
The Antarctic ice sheet is subject to extreme cold conditions, requiring a robot to be operational at  \SI{-40}{\celsius} for summer deployments, a temperature well below ratings of typical electronics~\citep{Hoffman2019}.
In combination with wind speed exceeding \SI{20}{\m \per \second}, cold will induce stress on the main body of a robotic platform~\citep{Hoffman2019}, affect manoeuvrability~\citep{Shillcutt1999}, crack seals relying on glue or tape letting snow inside~ \citep{Akers2004}, and any supporting equipment relying on small batteries will have a noticeable reduced lifespan~\citep{Stansbury2004}.
In terms of traversability, ice sheets and glaciers are equivalent to large plains subject to two specific hazards.
Firstly, hard snow will be eroded by winds creating waves-shaped ridges, sometimes referred to as sastrugi, which vary from a few centimetres to two metres high.
These obstacles can be high enough to flip a robot going downward or hard to overcome upward due to reduced traction.
Because of their shapes and lack of colours, sastrugi are notoriously hard to detect~\citep{Gifford2009}.
Moreover, these open areas have few structures limiting winds to gain speed, which will compact the snow and even carve blue-ice fields~\citep{Foessel1999}.
The second major hazard is large crevasses that can be hidden under thin layers of snow.
These crevasses threaten humans and vehicles navigating in the area.
Manual monitoring of these deep fractures is typically done using ground-penetrating radars.
Surveying hidden crevasses is another beneficial application for robots in these environments as it is dangerous and slow for humans to detect and mark their location~\citep{Trautmann2009,Lever2013}.
In terms of localization, the lack of vegetation produces large snow-covered plains, where the most useful information for visual-based localization is situated on the skyline~\citep{Barfoot2011}.
Having feature points on the horizon makes it more challenging to estimate the linear motion of a vehicle~\citep{Paton2017}.
Moreover, when navigating at a high latitude, the sun may move following the horizon for a long period.
This kind of motion of the sun cast long and rapidly moving shadows during the day, thus limiting the ability to relocalize in a given environment~\citep{Barfoot2010,Paton2017}.

Ice caps are also permanent ice anchored on the ground but cover a smaller area than ice sheets.
They can mostly be found on top of mountains, but are not limited to high elevation.
With the rapid development of \acp{UAV} combined with advances in photogrammetry, mountains and their snowpack can be surveyed at a faster pace.
However, \citet{Revuelto2021} observed that shadows cast by mountains impact greatly the 3D reconstruction, especially during winter, where the sun has lower incidence angles.
Overall, different sub-domains of the cryosphere are extremely challenging for robotic systems and must be well understood when expecting safe autonomous behaviours.


\subsection{Impact of precipitation on robotics}

\begin{figure*}[hbtp]
\begin{center}
\includegraphics[width=\textwidth,trim={0 2px 0 0},clip]{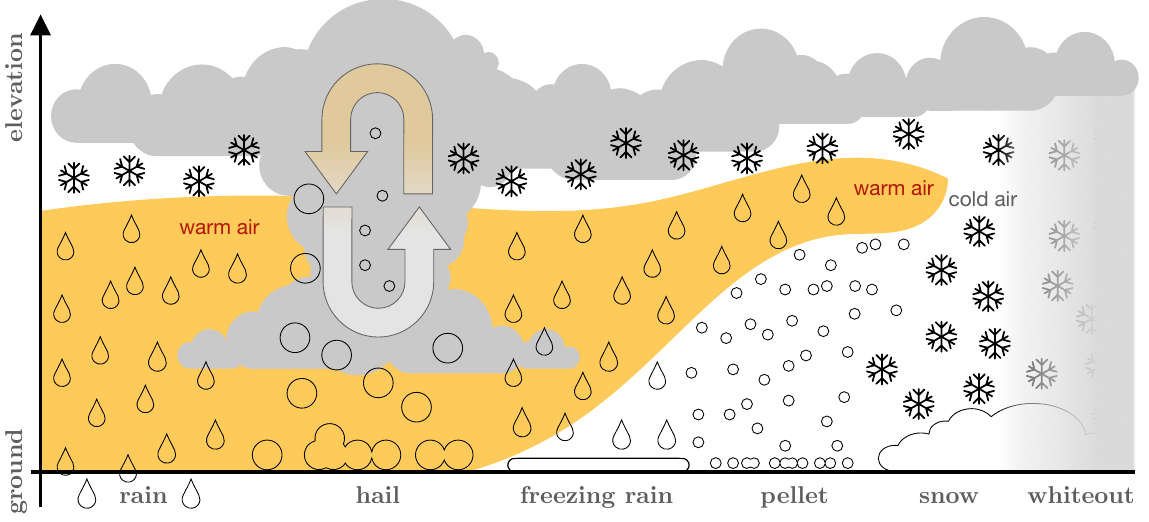}
\caption{Representation of the continuous transition between different types of precipitation from rain to snow.
Yellow represents warm air and white is cold air.
Different proportions of cold and warm air will make the transition between rain, hail, freezing rain, pellet, and snow.
While rain can be drained by the ground, solid precipitation will accumulate on the ground, thus changing rapidly the topology of the surroundings.
This dynamic topology combined with low visibility poses major challenges to robotic systems.}
\label{fig:precipitation}
\end{center}
\end{figure*}

The most famous type of precipitation associated with the cryosphere is snow.
In reality, the interaction of cold air and water is complicated and can produce a continuous spectrum of types of precipitation, such as the ones depicted in \autoref{fig:precipitation}.
%
In continental climates, it is common to see a mix of freezing rain, pellets, and snow during the same storm, especially close to spring and autumn, which can add challenges to perception algorithms. 

This variety of precipitation is an open problem for the localization of autonomous cars~\citep{Pitropov2020}.
Pellets and snow are known to degrade depth perception by adding noise to stereo and lidar sensing~\citep{Foessel1999}.
For example, a snowflake can trigger a lidar reading at up to \SI{20}{\m}~\citep{Charron2018}.
Millimetre-wave radar is less subjected to dense precipitation, but can still be occluded by wet snow sticking to the sensor optic~\citep{Hong2020}. 
Moreover, the potential size of hail can put the integrity of the robot at risk and can cause physical damage to light structures.
In general, freezing rain is known to be dangerous for infrastructure, which can collapse under the weight of the ice rapidly accumulating.
More related to robotics, a thick layer of ice can break antennas used for communication with a robot, attenuate \ac{GPS} signals used for localization, and hugely limit the efficiency of solar panels.
Moreover, the newly created slippery surfaces will perturb path tracking and control algorithms.

\begin{figure}[hbtp]
\begin{center}
\includegraphics[width=\columnwidth]{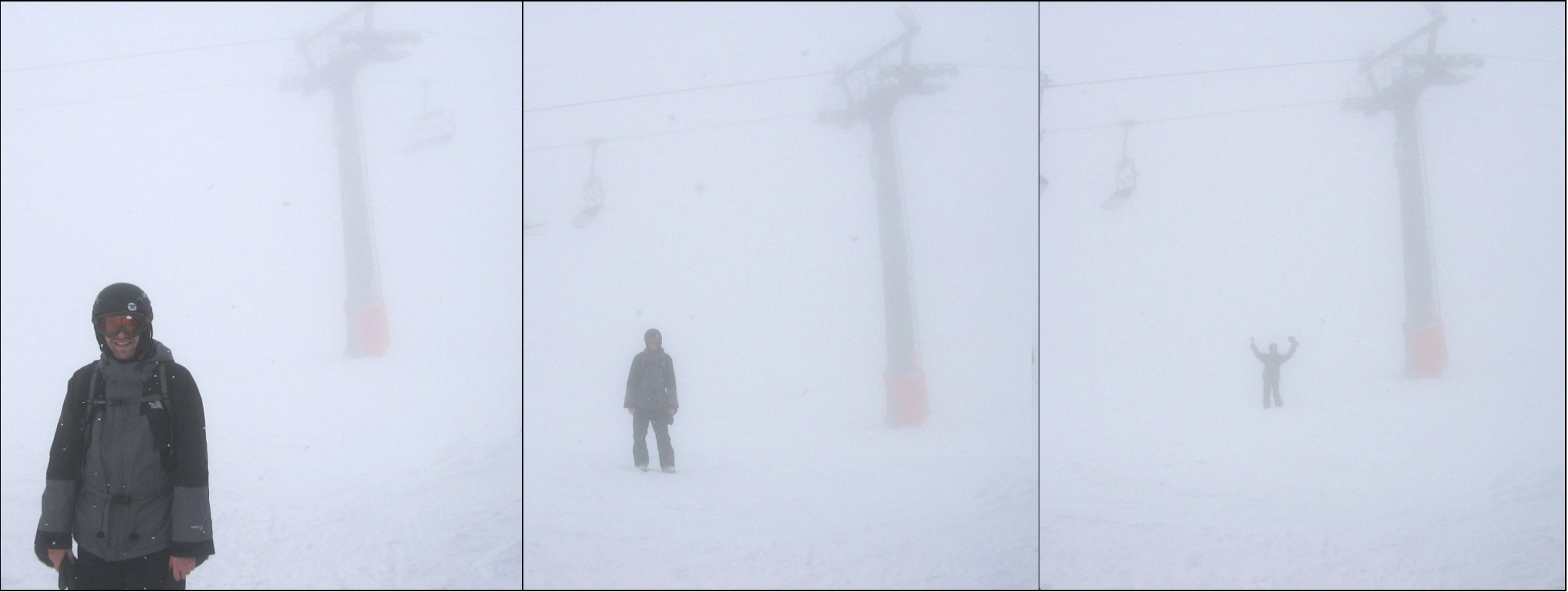}
\caption{Example of a perceptual challenge during a whiteout caused by fog and light snow on a mountain.
The lack of distinction between the ground and the sky combined with diffused lights pushes to the limit algorithms aiming at scene interpretation or localization. }
\label{fig:whiteout}
\end{center}
\end{figure}

Finally, whiteout conditions happen with a combination of snow and high wind or snow and fog.
Sustained high winds combined with snow are called a blizzard, during which it becomes uncertain whether the snow is rising from the ground or falling from the sky.
Throughout a whiteout, such as the one shown in \autoref{fig:whiteout}, light is completely diffused (i.e., not producing any shadows) rendering the ground and the sky indistinguishable.
These conditions reduce considerably the number of features that can be used to localize the robot.
It is not unusual to observe even humans feeling nauseous or subject to dysequilibrium when moving in a whiteout because of conflicting sensory stimulus~\citep{Hausler1995}.
Pilots will typically not fly in these conditions causing seriously delays~\citep{Williams2010} or even cancellations of robotics field deployments~\citep{Bonanno2003}.
Meteorological events, especially in cold regions, are diverse and have a direct impact on autonomous navigation safety.
Any long-term tasks will have to plan for precipitation sporadically limiting the perception of a robot in the cryosphere.

\section{Key Research Findings}

The current scientific literature is very sparse when it comes to field testing in ice and snow.
Nonetheless, pioneering works that helped gather critical field knowledge about deploying robots in harsh and cold conditions are highlighted here. 
Moreover, robots described in this section can also be found in \autoref{fig:zones}, showing how they spread over different sub-domains of the cryosphere.

In December 1992, an eight-legged robot named \emph{Dante} was deployed on the edge of an active volcano in Antarctica~\citep{Wettergreen1993}.
Although its descent was interrupted at an early stage, valuable lessons were learned when deploying this \SI{400}{\kg} walking robot on Mount Erebus and opened new research opportunities for remote exploration missions. 
Two years later, a new version named \emph{Dante~II} managed to rappel in another active volcano, Mount Spurr, Alaska~\citep{Bares1999}.
The robot could walk through a mix of snow, ice, rocks, and ashes to transmit images from the crater to volcanologists located in a remote control station.

Developed for more level surfaces, \emph{Nomad} was first validated on an ice shelf close to Patriot Hills Base Camp, Antarctica, in November 1998~\citep{Moorehead1999} before being sent to the periphery of a glacier of the same continent two years later~\citep{Apostolopoulos2000}.
Nomad drove \SI{10.3}{\km} in the harsh weather of Antarctica and could autonomously discover meteorites during its second deployment.

Often forgotten in the literature, modified \emph{Snowcats} using cameras as a guiding system was tested in the Italian Alps~\citep{Broggi2002}.
A Snowcat is a truck-sized vehicle equipped with a pair of tracks and is common on ski slopes to prepare the trails. 
In 2002, the modified vehicles were sent to Zucchelli Station, Antarctica, but the weather conditions were too critical to allow further testing~\citep{Bonanno2003}.
This unfortunate event highlights the challenge of field testing in remote sub-domains of the cryosphere. 

Moved solely by the wind, the \emph{Tumbleweed Polar Rover} was first tested on the ice sheet of Greenland~\citep{Behar2004} to be finally sent from the Amundsen-Scott South Pole Station, Antarctica, in 2004~\citep{JPL2004}.
The rover traveled around \SI{130}{\km} while transmitting information about its coordinates, temperature, and air pressure through a satellite link.

Following a similar validation sequence, \emph{MARVIN 1} started to experiment with polar conditions in the summer of 2003 in Greenland, before \emph{MARVIN 2} was deployed in Antarctica three years later~\citep{Gifford2009}.
In both cases, the robot demonstrated early potential for seismic and radar remote sensing of the ice sheet. 
Meanwhile, \emph{Cool Robot} and its upgraded version \emph{Yeti} were deployed from 2005 to 2011, also on both ice sheets.
These battery-powered robots traveled over \SI{600}{\km} in three consecutive field seasons.
This line of research culminated with deployment at the Old Pole, Antarctica, where ground-penetrating radar was used to survey the remains of an old camp~\citep{Williams2014}.

Finally, during the summer of 2009, a six-wheeled robot with an articulated chassis, named \emph{ROC6}, was deployed near the Haughton Crater, in the Canadian High Arctic~\citep{Barfoot2011}.
Although located at a very high latitude on Devon Island, this deployment was done on exposed permafrost, considered a polar desert, where ground-penetrating radar could be used to survey subsurface structures.

A handful of key observations are recurrent from these deployments.
Except for the Snowcats, all of these deployments were used to push the boundaries of robotics on one side and to validate extraterrestrial exploration solutions on the other.
Both ice sheets were confirmed to have interesting properties common to the frozen Europa moon~\cite{Gifford2009}, while polar deserts share similarities with the surface of Mars~\citep{Barfoot2011}.
Moreover, most of these deployments were done in large open spaces, where \ac{GPS} navigation was meeting the overall navigation accuracy of their missions.
Lastly, deploying a complex system in remote and harsh locations is costly and requires a high level of planning to fulfill scientific goals. 
Equipment needs to be shipped well in advance~\citep{Barfoot2010}, which puts pressure on the storage capability of fragile equipment in cold.
Moreover, the lack of infrastructure limits repair and modification to a minimum~\citep{Morad2020}.

\section{Examples of Application}

Beyond mockups for planetary exploration, robots being robust to snow and ice can support many applications.
Robotics can support Earth sciences in remote locations, where manual sampling of the environment can be tedious and sometimes dangerous.
Surveys using autonomous vehicles can bring higher spatiotemporal resolution and support many interdisciplinary sciences.
Past efforts were related to volcanology~\citep{Wettergreen1995}, mineralogy~\citep{Apostolopoulos2000}, atmospheric science~\citep{Behar2004}, seismology~\citep{Gifford2009}, glaciology~\citep{Ray2020}, and geomorphology~\citep{Barfoot2011} to name few.
Being able to move between two stations autonomously can support good transportation in remote locations.
Ice roads are often used to travel in winter, where building permanent infrastructure would be too expensive.
These roads, along with the main route between McMurdo Station and the Amundsen–Scott South Pole Station, need to be inspected regularly for crevasses or ice thickness.
Autonomous vehicles already demonstrated the potential to reduce risks associated with such tasks~\citep{Lever2013}.
Moreover, snow also needs to be regularly removed from roads to be accessible by less specialized vehicles.
Snow removal is an essential service in many inhabited areas with large snow cover and can be costly.
Snow removal is also a time-critical operation for many airports.
Also relying on frozen ground to avoid sinkage of the heavy machinery, forestry vehicles often operate during winter.
To sum up, applications related to snow and ice are not limited to expeditions in remote locations to show befits of automation.

\section{Future Directions for Research}

Self-driving cars are taking a larger place in the public sphere and are expected on our roads shortly. 
Although great technological advances have occurred during the last decade, autonomous driving during winter still raises reliability concerns.
More effort will be needed to make self-driving cars safe to use in continental climates and datasets on this topic are on the rise~\citep{Pavlov2019,Pitropov2020,Barnes2020}.
It is interesting to point out that, up to now, no autonomous navigation algorithm was demonstrated during whiteout conditions. 
Along with low visibility snowstorms, extreme meteorological events are difficult to plan for in a scientific experiment, thus making it difficult to validate the safety of robots in these conditions.
As it is a source of major scientific funding, missions preparing for space exploration are expected to continue~\citep{Reid2020}.
In the face of climate change, building finer models of the environment by using autonomous platforms to carry Earth science sensors in a harsh environment will also remain and even gain in importance.
Finally, roboticists will progress to extend autonomy solutions for different parts of the cryosphere.
For example, a research program named SNOW (Self-driving Navigation Optimized for Winter) investigates autonomous navigation solutions in subarctic forests \citep{Baril2020,Baril2022}.
Another interesting initiative is the 3D reconstruction of icebergs using the autonomous underwater vehicle \emph{Scolum}~\citep{Zhou2019}.
Finally, although early deployment in remote areas of the cryosphere relied on \acp{UGV}, \acp{UAV} were recently used to reconstruct snow-covered alpine mountains~\citep{Revuelto2021}.
Another interesting direction is the exploitation of the \emph{midnight sun} to extend the mission time of a solar-powered \ac{UAV} to monitor the calving of glaciers~\citep{Jouvet2018}.

At a higher level, four key challenges need to be addressed to enhance autonomy in snow and ice.
First, deployment in a remote location puts pressure on long-range communication in a context where there is little infrastructure to support deployment.
Second, energy economy or energy harvesting solutions need to be put forward for long-term autonomous missions. 
Currently, gas engines may interfere with environmental sampling, while battery performances degrade rapidly in cold.
Third, solutions for localization in rapidly changing environments and high precipitation need to be improved.
For example, snowbanks swiftly move with the wind, snow covers heavily deform under the weight of vehicles, and blizzard drastically degrades the perception capability of a robot.
Finally, mobility on ice and snow is notoriously difficult.
As an illustration, turning on the spot in deep snow will often stall a robot resulting in a vehicle digging itself even more as controllers try to compensate for an increasing tracking error. 
Moreover, path tracking algorithms need to cope with a larger spectrum of ground friction levels than what is typically tested in dry conditions.
At a micro level, the ground can be a variable composition of gravels, ice, and snow continuously changing the traction force of the robot.

Surely, the presented research trends combined with these ongoing challenges make \emph{robotics in snow and ice} an exciting research field for many years to come as potential applications evolve.
There is increasing demand to better understand rapid climate change, which will require higher spatiotemporal data to be collected in the cryosphere.
Moreover, there is rapidly growing investment in space ranging from planetary exploration to building habitats, bringing more attention to Earth testbeds.

\Urlmuskip=0mu plus 1mu\relax
\def\UrlBreaks{\do\/\do-}
\bibliographystyle{spbasic}  
\bibliography{references} 

\section{Cross-References}

\begin{itemize}
\item Navigation of Mobile Robots
\item Intelligent Vehicles
\item Space Robotics
\item Robotics in Forestry

\end{itemize}

\end{document}